%% file: root.tex
\documentclass[letterpaper, 10 pt, conference]{ieeeconf}  

\IEEEoverridecommandlockouts                              

\usepackage{graphicx,import} 
\usepackage{epsfig} 
\usepackage{amsmath} 
\usepackage{placeins}
\usepackage{color}

\title{
The Shortcomings of Force-from-Motion in Robot Learning
}

\author{Elie Aljalbout$^{1,2}$, Felix Frank$^{2}$, Patrick van der Smagt$^{2,3}$, and Alexandros Paraschos$^{2}$
\thanks{$^{1}$Elie Aljalbout is currently with the Robotics and Perception Group, at the Department of Informatics of the University of Zurich (UZH) and the Department of Neuroinformatics at UZH and ETH Zurich, Switzerland.}%
\thanks{$^{2}$During this work, all Authors were affiliated with the Machine Learning Research Lab at the Volkswagen Group, Munich, Germany.}%
\thanks{$^{3}$Department of Informatics, ELTE University Budapest.}%
}

\begin{document}

\maketitle
\thispagestyle{empty}
\pagestyle{empty}

\begin{abstract}
Robotic manipulation requires accurate motion and physical interaction control.
However, current robot learning approaches focus on motion-centric action spaces that do not explicitly give the policy control over the interaction.
In this paper, we discuss the repercussions of this choice and argue for more interaction-explicit action spaces in robot learning.
\end{abstract}

\section{Introduction}

Learning manipulation skills can be a key enabler for general-purpose robotics.
Recent work successfully demonstrated the ability of learning manipulation skills based on reinforcement and imitation learning~\cite{akkaya2019solving,alles2022learning,kim2024openvla}.
Initial efforts focused on learning control policies that act directly in the lowest-level of control of the robot~\cite{wahlstrom2015pixels}. 
Recently, however, in an effort to reduce the policy complexity and facilitate sim-to-real transfer~\cite{alles2022learning, martin2019variable, ulmer2021learning, allshire2021laser}, novel action spaces have been introduced to abstract the low level control particularities and platform-specific dependencies.
Action spaces are implemented as control feedback loops~\cite{alles2022learning, martin2019variable, ulmer2021learning}, motion primitives~\cite{bahl2020neural,aljalbout2021learning}, or latent action models~\cite{zhou_plas_2020,allshire2021laser, aljalbout2023clas}.
Their goal is to simplify the policy's role to outputting simpler commands such as position or velocity targets in either the task or configuration space of the robot.
In our recent work, we studied the effects of choosing an action space, different low-level feedback loops, and policy integration schemes on exploration, policy properties, and sim-to-real transfer~\cite{aljalbout2024role}.
Our results demonstrated that the choice of action space is \emph{crucial} for learning a policy in simulation and for its transfer to the real-world.

Action spaces that provide control over physical interactions have been proposed in~\cite{beltran2020learning,ulmer2021learning,martin2019variable, luo2019reinforcement}.
These action spaces are typically based on variable and adaptive impedance control or on force control in the low-level feedback loops.
However, motion-centric action spaces continue to be used for interaction tasks, despite their  limitations.
In this paper, we argue that such abstractions limit the policy's capability to perform certain manipulation tasks, we motivate the adoption of interaction-explicit representations, and we promote the design of suitable action spaces for general-purpose manipulation.

\section{Shortcomings of force-from-motion}
\label{sec:shortcomings}

\begin{figure}
    \centering
    \vspace*{0.2em}
    \hspace*{0.1em}
    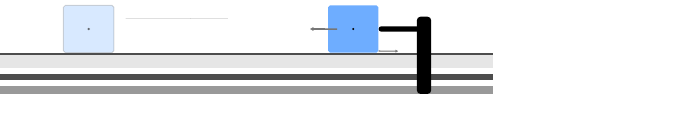
    \vspace*{-1.3em}
    \caption{%
    A 1-dimensional (1D) manipulation example.
    The task is to push the blue cube to the target. 
    The robot can move in the range $[q_\mathrm{min}, q_\mathrm{max}]$.
    The exerted force $F$, generated by the policy and the low-level controller, has to overcome the fiction $F_\mathrm{fric}$ for the cube to move.
    Using a motion-centric action space, such as joint positions, the robot can only apply forces indirectly by motion commands.
    That is by setting the low-level controller's target further away than the actual target position. 
    This approach has several shortcomings, as discussed in Sec.~\ref{sec:shortcomings} and 
    the use of an interaction-explicit action space overcomes them.
    \label{fig:d1_example}
    \vspace*{-1em}
    }
\end{figure}

Under \emph{force-from-motion}, manipulation policies can only implicitly exert forces onto their environment by overshooting their actual motion targets.
We illustrate the limitations of \emph{force-from-motion} in a simple 1D pushing example with a prismatic joint, as shown in Fig.~\ref{fig:d1_example}.
The policy outputs joint position targets and its goal is to move the blue cube to the target position.
The targets are tracked by a low-level joint impedance controller~\cite{hogan1985impedance}, that controls the joint force
\begin{equation}
    F = K (q_d - q) + D (\dot{q_d} - \dot{q}), 
    \label{eq:JIC}
\end{equation}
where $K$ and $D$ are the stiffness and damping gains, $q$ and $\dot{q}$ the current joint position and velocity, $q_d$ and $\dot{q_d}$ the joint position and velocity targets. 
Moving the cube requires applying a force with magnitude higher than $F_{\mathrm{min}}$, depending on the cube's mass and the surface friction properties. 
We set $D=0$ to simplify our discussion and we assume the robot force limits to be higher than $F_{\mathrm{min}}$.
The policy outputs are within the limits of the robot, i.e. $q_d \in [q_\mathrm{min}, q_\mathrm{max}]$.

For the policy to be able of moving the cube, we have
\begin{equation}
    F \ge F_\mathrm{min} \implies
    K \ge \frac{F_{\mathrm{min}}}{q_d - q},
\label{eq:constac}
\end{equation}
which is quite problematic for deciding the value of $K$ in the setup.
Setting, naively, $K = F_{\mathrm{min}}/(q_\mathrm{max} - q_\mathrm{min})$, the policy will only be able to push the cube at $q=q_\mathrm{min}$ by setting $q_d = q_\mathrm{max}$, or vice versa.
For any $q \in (q_\mathrm{min}, q_\mathrm{max})$, i.e.\ any position but the limits, the generated force $F$ is $F < F_\mathrm{min}$ and the robot will not be able to move the cube.
To increase the usable workspace, manipulate heavier objects, or if the contact surfaces have higher friction, $K$ needs to be increased.
Increasing $K$ soon becomes problematic, especially when the task requires the robot to be compliant or when a human is in the loop.
The constraint in Eq.~\eqref{eq:constac} renders the task unsolvable under those requirements.
Additionally, high values of $K$ and policy jitter can lead to force-clipping and unstable controllers.
Allowing the policy to set $q_d$ outside the physical limits of the joint $[q_\mathrm{min}, q_\mathrm{max}]$ can reduce $K$, but it leads to safety violations near the workspace limits and creates a trade-off between task feasibility and hardware safety.
Constraints similar to Eq.~\eqref{eq:constac} can also be derived for higher-order derivative action spaces.
As, typically, the magnitude of feasible velocities is greater than the range or the robot joints, using them allows for more compliant control due to the larger denominator.
Higher-order derivative action spaces are often adequate, despite not explicitly controlling the interaction forces~\cite{aljalbout2024role}.

Working with light objects, surfaces with low friction coefficients, and minimal human interaction alleviates \emph{force-from-motion}~\cite{akkaya2019solving,tang2023industreal} shortcomings.
Under these assumptions, these tasks can be successfully performed with almost any choice of action space~\cite{aljalbout2024role}. 

These shortcomings emerge as the \emph{force-from-motion} action spaces are not explicitly designed for interaction control and the policy applies forces indirectly from motion commands.
This illustrates how the choice of action space easily hinders task success.
While our 1D example is intentionally simplified, similar conclusions can be drawn for more general settings, e.g., for robots with more degrees of freedom.
Scaling robot learning to dynamic and human-robot interaction tasks would require more careful considerations.

\section{Discussion}
There are multiple approaches to overcome the shortcomings of the \emph{force-from-motion} action spaces.
Torque control, where the policy directly outputs joint-level torques, provides full control over the robot interactions. 
However, learning such policies in the real world is very challenging due to safety considerations. 
Training them first in simulation before deploying them on the real robot, while possible, suffers from a very large sim-to-real gap compared to other action spaces. 
This is due to the lack of feedback loops to compensate for dynamic mismatches between simulation and the real robot~\cite{aljalbout2024role}.
Delta action spaces, where the policy output is integrated to obtain a position or velocity target~\cite{tang2023industreal} provide a different approach on controlling the interaction forces, but have similar \emph{force-from-motion} limitations.
Delta action spaces introduce additional hidden dynamics and reduce the reactivity of the robot, that further degrade sim-to-real transfer~\cite{aljalbout2024role}.
Applied to our illustrative example, the robot will move the cube only after the position target is integrated sufficiently beyond, to generate the required force. 

To overcome the limitations of \emph{force-from-motion}, we can use interaction-explicit action spaces, as for example in~\cite{beltran2020learning,ulmer2021learning,martin2019variable, luo2019reinforcement}, or develop new ones.
Interaction-explicit action spaces can accurately control the interaction forces and are better suited for more dynamic manipulation tasks.
However, a notable drawback of these spaces is the difficulty of collecting data to train policies using imitation learning, which can significantly boost the learning process by training policies from demonstrations.
Interaction-explicit action spaces that are trainable from imitation are currently missing in the literature. 

In recent robot learning works, the \emph{force-from-motion} has been preferred for its simplicity and effectiveness in specific scenarios, particularly when manipulating light objects and the physical robot interactions are limited. 
However, it is inadequate for general-purpose robotics. 
This article demonstrates how \emph{force-from-motion} limits the range of learned behaviors and often results in undesirable effects (e.g., exceeding torque limits), even in basic scenarios. 
We have emphasized the necessity for more flexible action spaces that can better accommodate physical interactions and dynamic real-world tasks. 

Adopting interaction-explicit action spaces could mark a significant advancement towards more robust and general-purpose robotic manipulation learning.
Future work should further explore this direction and develop action spaces that are applicable to a large range of real-world-relevant manipulation tasks.

\bibliographystyle{ieeetr}
\bibliography{biblio}

\end{document}

%% file: figures/rect1.pdf_tex
\begingroup%
  \makeatletter%
  \providecommand\color[2][]{%
    \errmessage{(Inkscape) Color is used for the text in Inkscape, but the package 'color.sty' is not loaded}%
    \renewcommand\color[2][]{}%
  }%
  \providecommand\transparent[1]{%
    \errmessage{(Inkscape) Transparency is used (non-zero) for the text in Inkscape, but the package 'transparent.sty' is not loaded}%
    \renewcommand\transparent[1]{}%
  }%
  \providecommand\rotatebox[2]{#2}%
  \newcommand*\fsize{\dimexpr\f@size pt\relax}%
  \newcommand*\lineheight[1]{\fontsize{\fsize}{#1\fsize}\selectfont}%
  \ifx\svgwidth\undefined%
    \setlength{\unitlength}{333.73845859bp}%
    \ifx\svgscale\undefined%
      \relax%
    \else%
      \setlength{\unitlength}{\unitlength * \real{\svgscale}}%
    \fi%
  \else%
    \setlength{\unitlength}{\svgwidth}%
  \fi%
  \global\let\svgwidth\undefined%
  \global\let\svgscale\undefined%
  \makeatother%
  \begin{picture}(1,0.17132114)%
    \lineheight{1}%
    \setlength\tabcolsep{0pt}%
    \put(0,0){\includegraphics[width=\unitlength,page=1]{rect1.pdf}}%
    \put(0.44111555,0.13953178){\color[rgb]{0,0,0}\makebox(0,0)[lt]{\lineheight{1.25}\smash{\begin{tabular}[t]{l}\scriptsize $F$\end{tabular}}}}%
    \put(0.54754169,0.10532872){\color[rgb]{0,0,0}\makebox(0,0)[lt]{\lineheight{1.25}\smash{\begin{tabular}[t]{l}\scriptsize $F_\mathrm{fric}$\end{tabular}}}}%
    \put(0.17187501,0.15083237){\color[rgb]{0,0,0}\makebox(0,0)[lt]{\lineheight{1.25}\smash{\begin{tabular}[t]{l}\scriptsize target\end{tabular}}}}%
    \put(0.0175745,0.00542506){\color[rgb]{0,0,0}\makebox(0,0)[lt]{\lineheight{1.25}\smash{\begin{tabular}[t]{l}\small $q_\mathrm{max}$\end{tabular}}}}%
    \put(0.63560008,0.00542506){\color[rgb]{0,0,0}\makebox(0,0)[lt]{\lineheight{1.25}\smash{\begin{tabular}[t]{l}\small $q_\mathrm{min}$\end{tabular}}}}%
  \end{picture}%
\endgroup%